%% file: main.tex
\documentclass[10pt,twocolumn,letterpaper]{article}

\usepackage{cvpr}              

\usepackage{graphicx}
\usepackage{amsmath}
\usepackage{amssymb}
\usepackage{booktabs}
\usepackage{xcolor}
\usepackage{kotex}
\usepackage{multirow}
\usepackage{balance}
\usepackage[linesnumbered,lined,ruled,noend]{algorithm2e}

\usepackage[pagebackref,breaklinks,colorlinks]{hyperref}

\usepackage[capitalize]{cleveref}
\crefname{section}{Sec.}{Secs.}
\Crefname{section}{Section}{Sections}
\Crefname{table}{Table}{Tables}
\crefname{table}{Tab.}{Tabs.}

\def\cvprPaperID{116} 
\def\confName{CVPR}
\def\confYear{2022}

\begin{document}

\title{Demystifying the Neural Tangent Kernel from a Practical Perspective: \\ Can it be trusted for Neural Architecture Search without training?}
\author{Jisoo Mok$^1$\thanks{Work done while interning at NAVER (magicshop1118@snu.ac.kr)} ~~~~~~~ Byunggook Na$^1$ ~~~~~~~ Ji-Hoon Kim$^{2,3\dagger}$ ~~~~~~ Dongyoon Han$^{2\dagger}$ ~~~~~~~  Sungroh Yoon$^{1,4}\thanks{Corresponding Authors}$\\
$^1$ Department of ECE, Seoul National University ~~~~~ $^2$ NAVER AI Lab ~~~~~ $^3$ NAVER CLOVA\\
$^4$ AIIS, ASRI, INMC, ISRC, and Interdisciplinary Program in AI, Seoul National University}
  
\maketitle

  
\begin{abstract}
In Neural Architecture Search (NAS), reducing the cost of architecture evaluation remains one of the most crucial challenges.
Among a plethora of efforts to bypass training of each candidate architecture to convergence for evaluation, the Neural Tangent Kernel (NTK) is emerging as a promising theoretical framework that can be utilized to estimate the performance of a neural architecture at initialization.
In this work, we revisit several at-initialization metrics that can be derived from the NTK and reveal their key shortcomings.
Then, through the empirical analysis of the time evolution of NTK, we deduce that modern neural architectures exhibit highly non-linear characteristics, making the NTK-based metrics incapable of reliably estimating the performance of an architecture without some amount of training.
To take such non-linear characteristics into account, we introduce Label-Gradient Alignment (LGA), a novel NTK-based metric whose inherent formulation allows it to capture the large amount of non-linear advantage present in modern neural architectures.
With minimal amount of training, LGA obtains a meaningful level of rank correlation with the final test accuracy of an architecture.
Lastly, we demonstrate that LGA, complemented with few epochs of training, successfully guides existing search algorithms to achieve competitive search performances with significantly less search cost. 
The code is available at: \url{https://github.com/nutellamok/DemystifyingNTK}.
\end{abstract}

\section{Introduction}
\label{sec:intro}
Deep Neural Networks (DNNs) continue to produce impressive results in a wide variety of domains and applications.
The remarkable success of DNNs is due in no small part to the development of novel neural architectures, all of which used to be designed manually by machine learning engineers by testing a number of architectural design choices.
To remedy this issue, Neural Architecture Search (NAS), a sub-field of automated machine learning, has emerged as a feasible alternative to hand-designing neural architectures~\cite{elsken2019neural}.

\begin{figure}[t]
\begin{center}
   \includegraphics[width=\linewidth]{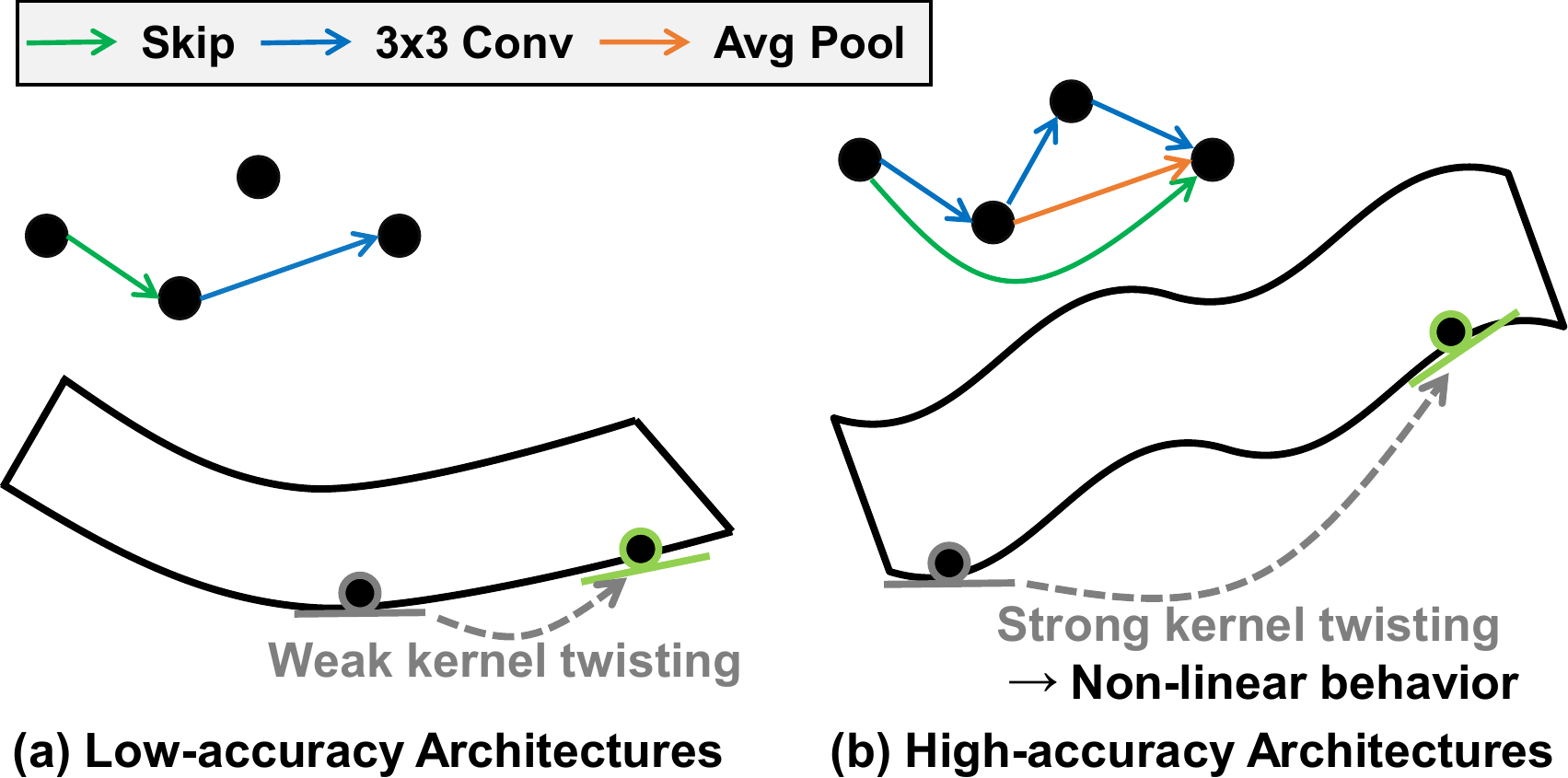}
\end{center}
\vspace{-15pt}
   \caption{ A conceptualized view of how (a) the NTK of a low-accuracy architecture and (b) that of a high-accuracy architecture evolve during training (gray $\rightarrow$ green). Black planes denote the function space realization of weight parameters. On the top left corners of (a) and (b), an example of a low- and high-accuracy architecture is provided. Unlike a low-accuracy architecture, a high-accuracy architecture equipped with a large amount of non-linear advantage experiences strong kernel twisting, such that the principal components of the NTK become more aligned with target labels. In~\figurename~\ref{fig:metric_bench_201_all}, we illustrate that LGA captures this difference in the architectures residing in two polar accuracy regimes.}
\label{fig:acc_rob_curve}
\vspace{-10pt}
\end{figure}

Although the architectures derived by NAS are beginning to outperform hand-designed architectures, the tremendous computational cost required to execute NAS makes its immediate deployment rather challenging~\cite{real2017large, le2017naswithRL, zoph2018nasnet}.
The majority of the search cost in NAS is induced by the need to train each candidate architecture to convergence for evaluation~\cite{dean2018enas}.
In more recently proposed NAS algorithms, the individual training of candidate architectures is circumvented by a weight-sharing strategy~\cite{dean2018enas, yang2019darts, xu2019pc, chu2019fairdarts, zhou2020NAS, chen2020drnas, dong2019searching, peng2020cream, zhang2020overcoming}.
With weight sharing, the computational cost of NAS is reduced by orders of magnitude: from tens of thousands of GPU hours to $<1$ GPU day.
Unfortunately, NAS algorithms that rely on weight-sharing experience an optimization gap between the performance of an architecture approximated through weight-sharing and its stand-alone performance~\cite{xie2020weight}.

Another line of research that aims to accelerate the architecture evaluation process focuses on developing a performance predictor with as few architecture-accuracy pairs as possible.
Minimizing the mean squared error between the predicted and ground-truth accuracy is the most straightforward way of training such a performance predictor because the problem of performance prediction can naturally be interpreted as a regression task~\cite{li2020gp, deng2017peephole, tang2020semi}.
The family of architecture comparators replaces the deterministic evaluation of neural architectures with a relativistic approach that compares two architectures and determines which one yields better performance~\cite{xu2021renas, chen2021contrastive}. 
Apart from weight sharing and performance prediction, some works propose more general proxies for architecture evaluation~\cite{na2021accelerating, zhou2020econas, lin2021zen, mellor2021neural, abdelfattah2020zero}.
White \textit{et al.}~\cite{white2021powerful} offer a comprehensive survey of performance predictors in NAS, and in the Appendix, we discuss related works in more detail.

The need to explicitly measure the test accuracy of architectures or train a performance predictor arises from our lack of theoretical understanding regarding how and what DNNs learn.
Among diverse deep learning theories that claim to offer a quantifiable bound on the learning capacity of DNNs, the Neural Tangent Kernel (NTK) framework~\cite{jacot2018neural} is garnering a particular amount of attention. 
Based on the observation that DNNs of infinite widths are equivalent to Gaussian processes, the NTK framework proposes to characterize DNNs as kernel machines~\cite{jacot2018neural}.
At the core of the NTK framework lies an assumption that the NTK computed from an infinitely wide DNN parameterized with randomly initialized weights remains unchanged throughout training.
Thus, the NTK framework suggests that the training dynamics of such a DNN can be fully characterized by the NTK at initialization. 
Motivated by the solid theoretical ground on which the NTK framework is built, in the field of NAS, NTK-based metrics~\cite{xu2021knas, chen2020neural}, measured at initialization, have been proposed as an attractive alternative to computing the test accuracy directly.

In this paper, we aim to rigorously evaluate how trustworthy of a theory the NTK framework is in the context of NAS by conducting a series of empirical investigations.
To begin with, we revisit previously-proposed metrics that spawn out of the NTK framework: Frobenius Norm (F-Norm), Mean~\cite{xu2021knas}, and Negative Condition Number (NCN)~\cite{chen2020neural}.
In order to assess whether the NTK-based metrics, computed with randomly initialized weights, are truly applicable to NAS, we test them on various NAS benchmarks by measuring Kendall's Tau rank correlation with the test accuracy at convergence.
Our experimental results show that the predictive performance of the NTK-based metrics obtained at initialization fluctuates significantly from one benchmark to another.
A more up-close study of how their predictive ability changes according to the evaluated architecture pool and the weight initialization scheme uncovers additional pitfalls of the NTK framework.
Comprehensively, our results seem to indicate that the NTK at initialization does not exhibit a substantial level of reliability for architecture selection.

\begin{figure}[t]
\begin{center}
   \includegraphics[width=0.9\linewidth]{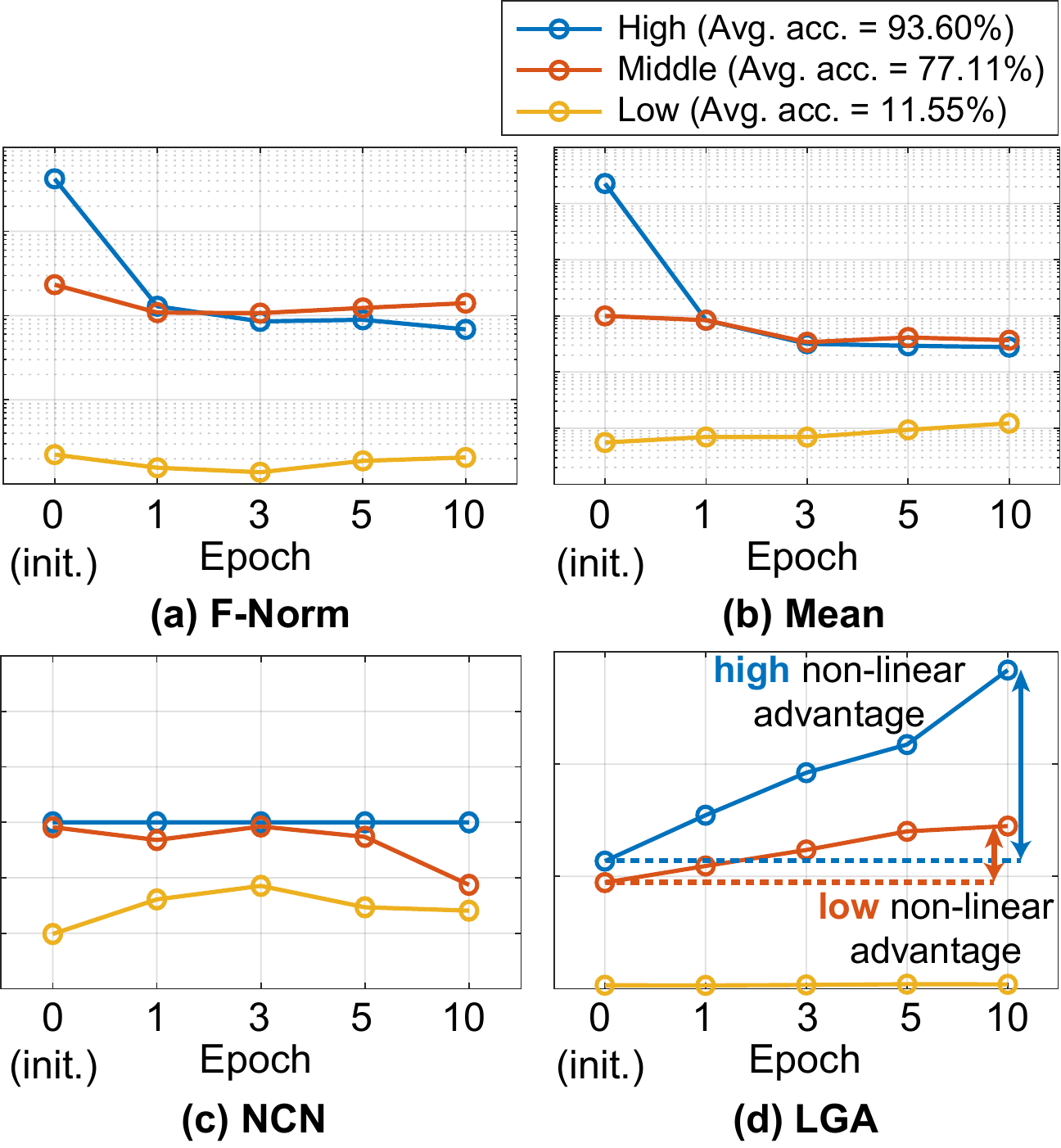}
\end{center}
\vspace{-10pt}
   \caption{ How the NTK-based metrics change in the early epochs for high-, mid-, and low-accuracy architectures. For each accuracy range, 200 architectures are randomly sampled from the NAS-Bench-201 search space, and the averaged test accuracy per architecture set is included in the legend.}
\label{fig:metric_bench_201_all} 
\vspace{-20pt}
\end{figure}

Empirically analyzing the time evolution of the NTK reveals that in modern neural architectures that constitute NAS search spaces, the NTK evolves in a highly non-linear manner.
As a result, modern neural architectures tend to exhibit a large amount of non-linear advantage~\cite{fort2020deep, goldblum2019truth, ortiz2021can}.
~\figurename~\ref{fig:acc_rob_curve} depicts on a high-level how the NTK rotates and evolves during the training process.
Inspired by this observation, we introduce Label-Gradient Alignment (LGA), a novel NTK-based metric whose mathematical formulation allows it to coherently capture the non-linear characteristics of modern neural architectures.
After only few epochs of training, LGA shows a considerable level of rank correlation with the test accuracy at convergence.
As illustrated in~\figurename~\ref{fig:metric_bench_201_all}, delving deeper into how each metric changes throughout training consolidates that LGA is the only metric that can accurately reflect the non-linear behavior of modern neural architectures.
Lastly, we conduct random~\cite{li2020random} and evolutionary search~\cite{real2019regularized} algorithms solely by using post-training LGA to demonstrate that it can be used to accelerate existing search algorithms.

Our main contributions can be summarized as follows:
\begin{itemize}
\item{We rigorously assess the predictive ability of previous NTK-based metrics on various NAS benchmarks and under different hyperparameter settings. Our results imply that the NTK at initialization may be insufficient for architecture selection in NAS.}
\item{In order to understand the cause of the aforementioned limitation of the current NTK framework, we analyze the time evolution of the NTK and reveal that a considerable amount of non-linear advantage is present in modern neural architectures considered in NAS.}
\item{We introduce LGA, a novel NTK-based metric that can reflect the change in NTK with respect to the target function. Integrating LGA after minimal amount of training with existing search algorithms yields a competitive search performance to state-of-the-art NAS algorithms, while noticeably reducing the search cost.}
\end{itemize}

\section{Neural Tangent Kernel}
This section provides an overview of the NTK framework and the NTK-metrics that will be subject to a series of investigations in the later sections.
In Section~\ref{prelim}, we introduce the concept of the NTK, and in Section~\ref{ntk_metric}, we briefly review previously proposed NTK-based metrics.

\subsection{Preliminaries}\label{prelim}
Let us define a DNN as a function $f_{\theta}: \mathbb{R}^{d} \rightarrow \mathbb{R}$, where $\theta$ is the set of trainable weight parameters. 
Given the target dataset, $\mathcal{D} = \{(x_i, y_i)\}_{i=1}^{N}$, without loss of generality, the NTK framework focuses on a binary classification problem, whose objective is to minimize the squared loss, $L_{\mathcal{D}}(\theta) = \sum_{i=1}^{N} ||y_i - f_\theta(x_i)||_2^2$.
Here, $x_i \in \mathcal{X}$ and $y_i \in \mathcal{Y}$ denote image samples and the corresponding class labels, respectively.
In a small neighborhood region around the randomly initialized weights $\theta_0$, a DNN can be linearly approximated through a first-order Taylor expansion~\cite{jacot2018neural}:
\begin{equation}\label{linear_approx}
    f_\theta(x) \approx \hat{f_\theta}(x; \theta) = f_{\theta_0}(x) + (\theta - \theta_0)^\intercal \nabla_\theta f_{\theta_0}(x),
\end{equation}
where $\nabla_\theta f_{\theta_0}$ corresponds to the Jacobian of a DNN's prediction, computed with respect to $\theta_0$. 
The obtained approximation $\hat{f_\theta}$ can be regarded as a linearized network that maps weight vectors to functions residing in a reproducible kernel Hilbert space (RKHS) $\mathcal{H} \subseteq L_2(\mathbb{R}^d)$, determined by the NTK at $\theta_0$~\cite{ortiz2021can}: 
\begin{equation}
    \Theta_{\theta_0}(x, x') = \langle \nabla_\theta f_{\theta_0}(x) , \nabla_\theta f_{\theta_0} (x')^\intercal \rangle.
\end{equation}
Note that the NTK is essentially the dot product of two gradient vectors, and is thus equivalent to the Gram matrix of per-sample gradients.
Intuitively speaking, the NTK can be interpreted as a condensed representation of gradient values and gradient correlations.
From a geometric perspective, gradient values influence the extent of gradient descent at each step, and gradient correlations determine the stochasticity of gradient directions~\cite{xu2021knas}.

It has recently been discovered that under an infinitesimal learning rate and certain types of initialization, in a DNN of infinite width, the approximation in Eq.~(\ref{linear_approx}) is exact, and the NTK remains constant throughout training.
Therefore, provided that the above assumptions hold, several aspects of the NTK at initialization can be used to fully characterize the training dynamics of a DNN and estimate its generalization performance.
Following this theoretical discovery, in NAS, Xu~\textit{et al.}~\cite{xu2021knas} and Chen~\textit{et al.}~\cite{chen2020neural} have proposed to score DNNs at initialization based on the metrics that spawn out of the NTK.

\subsection{Previous NTK-based Metrics}\label{ntk_metric}
\noindent\textbf{Metric I: Frobenius Norm} 
Suppose $\Theta_{\theta_t}$ is the NTK at the $t$-th epoch. According to Xu~\textit{et al.}~\cite{xu2021knas}, for any $t > 0$, the following inequality holds:
\begin{equation}\label{fnorm_proof}
    ||y_i-f_{\theta_t}(x_i)||_2^2
    \leq exp(-\lambda_\mathrm{min}t)||y_i-f_{\theta_0}(x_i)||_2^2,
\end{equation}
where $\lambda_\mathrm{min}$ is the minimum eigenvalue of the NTK matrix $\Theta_{\theta_t}$.
From Eq.~(\ref{fnorm_proof}), we see that the upper bound on the loss term is determined by $\lambda_\mathrm{min}$; the larger $\lambda_\mathrm{min}$ is, the tighter the upper bound becomes, thereby yielding a smaller training loss.

Since $\Theta_{\theta_t}$ is always symmetrical by definition, $\lambda_\mathrm{min}$ can be bounded by the Frobenius norm of $\Theta_{\theta_t}$:
\begin{equation}
    \lambda_\mathrm{min} \leq \sqrt{\sum_k |\lambda_k|^2} = ||\Theta_{\theta_t}||_F,
\end{equation}
where $\lambda_k$ denotes the $k$-th eigenvalue of $\Theta_{\theta_t}$, ordered by $\lambda_\mathrm{{min}} \leq \ldots \leq \lambda_{\mathrm{max}}$.
Utilizing the Frobenius norm as a metric to score DNNs allows us to circumvent the eigendecomposition of $\Theta_{\theta_t}$ with the time complexity of $\mathcal{O}(n^3)$.
Provided that the NTK does remain constant regardless of training as mentioned in Section~\ref{prelim}, for any value of $t$, $||\Theta_{\theta_t}||_F$ can be replaced with $||\Theta_{\theta_0}||_F$.
For the remainder of this paper, we use the abbreviation \textit{F-Norm} to refer to this metric, which must be \textit{positively correlated} with the final test accuracy of a DNN.\\
\noindent\textbf{Metric II: Mean} 
Although Xu~\textit{et al.}~\cite{xu2021knas} show that the $||\Theta_{\theta_0}||_F$ can be leveraged to evaluate randomly-initialized DNNs, they do not directly use F-norm as a metric. 
Instead, the mean of $\Theta_{\theta_0}$ is proposed as a metric for evaluating DNNs at initialization.
The mean of the NTK matrix, denoted by $\mu({\Theta}_{\theta_0})$, can be expressed as follows:
\begin{equation}
\mu({\Theta}_{\theta_0}) = \frac{1}{N^2}\sum_{i=1}^{N}\sum_{j=1}^{N} \left(\frac{\partial f_{\theta_0}(x_i)}{\partial \theta_0} \right)\left(\frac{\partial f_{\theta_0}(x_j)}{\partial \theta_0} \right)^{\intercal}
\end{equation}
Like F-norm, the Mean metric must also be \textit{positively correlated} with the final test accuracy.

\noindent\textbf{Metric III: Negative Condition Number} Lee \textit{et al.}~\cite{lee2019wide} prove that the training dynamics of infinitely wide DNNs are controlled by ordinary differential equations that can be solved as:
\begin{equation}\label{cn_1}
    f_{\theta_{t}}(\mathcal{X}) = (\mathbf{I} - exp(-\eta \Theta_{\theta_t} t)) \mathcal{Y},
\end{equation}
where $\eta$ and $\mathbf{I}$ represent the learning rate and the Identity matrix, respectively.
Lee~\textit{et al.} also hypothesize that the maximum feasible learning rate is given by: $\eta \sim 2\slash \lambda_{\mathrm{max}}$.
A further study into the relationship between $\Theta_{\theta_t}$ and the trainability of DNNs leads Xiao~\textit{et al.}~\cite{xiao2019disentangling} to conclude that Eq.~(\ref{cn_1}) can be re-written in terms of the eigenspectrum of $\Theta_{\theta_t}$ as follows:
\begin{equation}\label{cn_2}
    f_{\theta_{t}}(\mathcal{X}) = (\mathbf{I} - exp(-\eta \lambda_k t)) \mathcal{Y},
\end{equation}
where $\lambda_k$ denotes the $k$-th eigenvalue of $\Theta_{\theta_t}$.
Plugging the maximum feasible learning rate $2\slash \lambda_{\mathrm{min}}$ into Eq.~(\ref{cn_2}), Chen~\textit{et al.}~\cite{chen2020neural} see that $\lambda_{\mathrm{min}}$ converges exponentially at the rate of $1 \slash c$, where $c = \lambda_{\mathrm{max}} \slash \lambda_{\mathrm{min}}$ is the condition number (CN) of $\Theta_{\theta_t}$.
As CN grows larger, the output of a DNN $f_{\theta_{t}}(\mathcal{X})$ will fail to converge to the target label $\mathcal{Y}$.
Thus, CN must exhibit a negative correlation with the final test accuracy.
In this paper, to keep the trend in rank correlation consistent with the rest of the investigated metrics, we use the Negative Condition Number (NCN) instead; hence, NCN must be \textit{positively correlated} with the final test accuracy.



\section{Limitations of the NTK at Initialization}\label{sec:3}
Here, we test the universal applicability of previous NTK-based metrics, measured at initialization, to diverse search spaces offered by NAS benchmarks.
Even though these at-initialization metrics have been believed to highly correlated with the final accuracy, empirical demonstrations of their predictive abilities have been limited to a single search space: NAS-Bench-201.
Therefore, extending the scope of evaluation to a far more diverse set of search spaces that contain different candidate operations and connectivity patterns is crucial for rigorously verifying the reliability of the NTK-based metrics.
In Section~\ref{benchmark}, we provide a summary of the NAS benchmarks that are used for evaluation; more details on the construction of these benchmarks, as well as the image datasets they utilize, can be found in the Appendix.
In Section~\ref{benchmark_results}, we present the evaluation results and report key findings regarding the practicality of the NTK-based metrics.
Lastly, in Sections~\ref{sec:3_3} and~\ref{sec:3_4}, we discuss additional pitfalls of the NTK identified from more up-close analyses of the NTK-based metrics.

\subsection{Benchmarks for Neural Architecture Search}\label{benchmark}
\noindent\textbf{NAS-Bench-101}~\cite{ying2019bench} contains 423,000 computationally unique neural architectures evaluated on CIFAR-10~\cite{krizhevsky2009learning}.
All of the architectures in NAS-Bench-101 adopt the cell topology, a smaller feedforward module that is stacked repeatedly to construct the final architecture.
The maximum depth of each cell and the number of possible connections are set to be 7 and 9, respectively, and the following are the available candidate operations: $3 \times 3$ convolution, $1 \times 1$ convolution, and $3 \times 3$ max pooling.\\
\textbf{NAS-Bench-201}~\cite{dong2019bench} contains 15,625 architectures, all of which are evaluated on CIFAR-10, CIFAR-100~\cite{krizhevsky2009learning}, and ImageNet-16-120~\cite{chrabaszcz2017downsampled}.
Similar to NAS-Bench-101, NAS-Bench-201 architectures are also based on the cell topology.
Each one of the cells in NAS-Bench-201 has the fixed depth of 4, and the following candidate operations are included in the search space: zeroize, skip connection, $1 \times 1$ convolution, $3 \times 3$ convolution, and $3 \times 3$ average pooling.\\
\textbf{NDS}~\cite{radosavovic2019network} offers a comprehensive analysis of commonly-adopted search spaces in NAS. 
The search spaces supported by the NDS benchmark include: DARTS~\cite{yang2019darts}, ENAS~\cite{dean2018enas}, NASNet~\cite{zoph2018nasnet}, AmoebaNet~\cite{real2019regularized}, and PNAS~\cite{liu2018progressive}. 
Although all of these search spaces adopt the cell topology, the design of the cell structure differs from one another; please refer to the Appendix for the summary of differences.
How the cells are stacked to generate the final neural architecture also varies among papers, but NDS standardizes this aspect of the search space by utilizing the DARTS architecture configuration.
For each search space, NDS trains and evaluates $\sim 1K$ architectures on CIFAR-10. 

\begin{figure}[t]
\begin{center}
   \includegraphics[width=\linewidth]{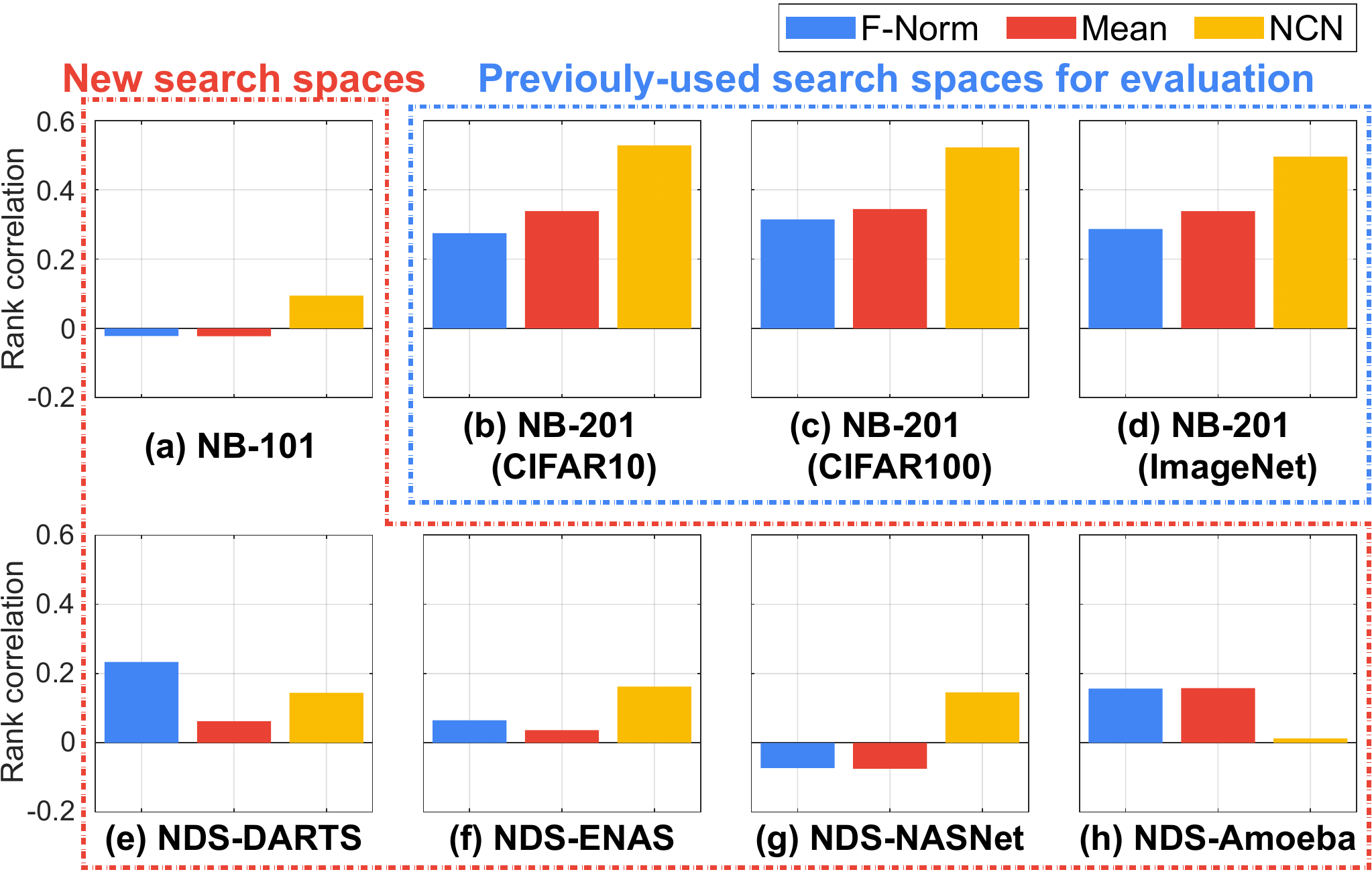}
\end{center}
\vspace{-10pt}
   \caption{ Rank correlation evaluation results on various NAS benchmarks. We compute the three metrics using Train and Eval Mode BNs. For simplicity, the higher correlation coefficient obtained from the two settings is reported here. The scale and the range of y-axes are set to be the same across all search spaces.}
\label{fig:nas_bench_all}
\vspace{-10pt}
\end{figure}

\subsection{Benchmark Evaluation Results}\label{benchmark_results}
By using Kendall's Tau as the measure of rank correlation, we evaluate how reliably the NTK-based metrics computed at initialization can predict the final test accuracy on various search spaces.
For the sake of computational efficiency, we randomly sample 1,000 architectures from each search space for evaluation.
It also appears that there exists no consensus on which batch statistics must be used for the batch normalization (BN) layer~\cite{ioffe2015batch} when computing the NTK. 
We thus test out both Train and Eval mode BN available in PyTorch~\cite{paszke2019pytorch}.
Please refer to the Appendix for detailed experimental settings used in this section.

In~\figurename~\ref{fig:nas_bench_all}, we report the abbreviated evaluation results that only include the highest rank correlation coefficient obtained for each metric; a comprehensive visualization of all rank correlation measurement results is provided in the Appendix.
Due to the page constraint, the results on NDS-PDARTS have also been moved to the Appendix.
On NAS-Bench-201, we have successfully reproduced the rank correlation measure for Mean~\cite{xu2021knas} and NCN~\cite{chen2020neural} reported in their original papers.
In NAS-Bench-101 and NDS search spaces, the degree of rank correlation decreases noticeably for all three metrics.
In NAS-Bench-101 and NDS-NASNet, in particular, F-Norm and Mean seem to be negatively correlated with the final test accuracy, which goes against their theoretical motivation.
Considering that the search spaces of NAS-Bench-101 and NDS are more complicated than that of NAS-Bench-201, such results may call into question whether the NTK framework can be deployed universally to more complex search spaces.

We also note that no single BN usage seems compatible with all three metrics.
For instance, on the one hand, using fixed batch statistics (\ie Eval mode BN) generally yields high rank correlation for NCN in NAS-Bench-201.
On the other hand, in the same search space, using per-sample batch statistics (\ie Train mode BN) improves rank correlation for F-Norm and Mean.
This finding makes it evident that the NTK framework as is may lack consideration of the effect of BN on modern neural architectures.




\subsection{Fine-grained Rank Correlation Evaluation}\label{sec:3_3}
In the previous section, 1,000 architectures were randomly sampled from each benchmark to uniformly represent the entire architecture set.
We now design a more challenging experiment, where we rank architectures in descending order and divide them into deciles, denoted by P; P1 contains Top-$10\%$ of architectures, P2 contains Top-$10\sim20\%$ of architectures, and so on.
From each decile, 100 architectures are sampled for evaluation.
This experiment allows us to determine whether the NTK-based metrics can stably guide the search process by gradually searching for a better architecture.
Such a fine-grained experiment is no longer valid in search spaces that contradict the theoretical motivation.
Therefore, the experiments in this section are conducted only on NAS-Bench-201.
 
We repeat this experiment with 20 different seeds for architecture sampling and visualize the results in the form of box-and-whisker plots. Please refer to Section A6 and Figures A2, A3, and A4 in the Appendix for the evaluation results. They suggest that in most deciles, the predictive ability of NTK-based metrics fluctuates significantly according to the choice of architectures used for evaluation.
These results imply that guiding a search algorithm with the NTK-based metrics may not be able to escape from a locally optimal architecture and thus may often lead to unstable search results.
Also, progressively shrinking the initial search space based on the error distribution of the architectures within it has become a commonly adopted technique in NAS or general architecture design~\cite{li2020sgas,radosavovic2020designing,hu2020angle,chen2020drnas}.
In such a refined search space that consists only of high-accuracy architectures, NTK-based metrics may fail to identify a particularly better architecture.



\subsection{Sensitivity to Weight Initialization}\label{sec:3_4}
\begin{figure}[t]
\begin{center}
   \includegraphics[width=\linewidth]{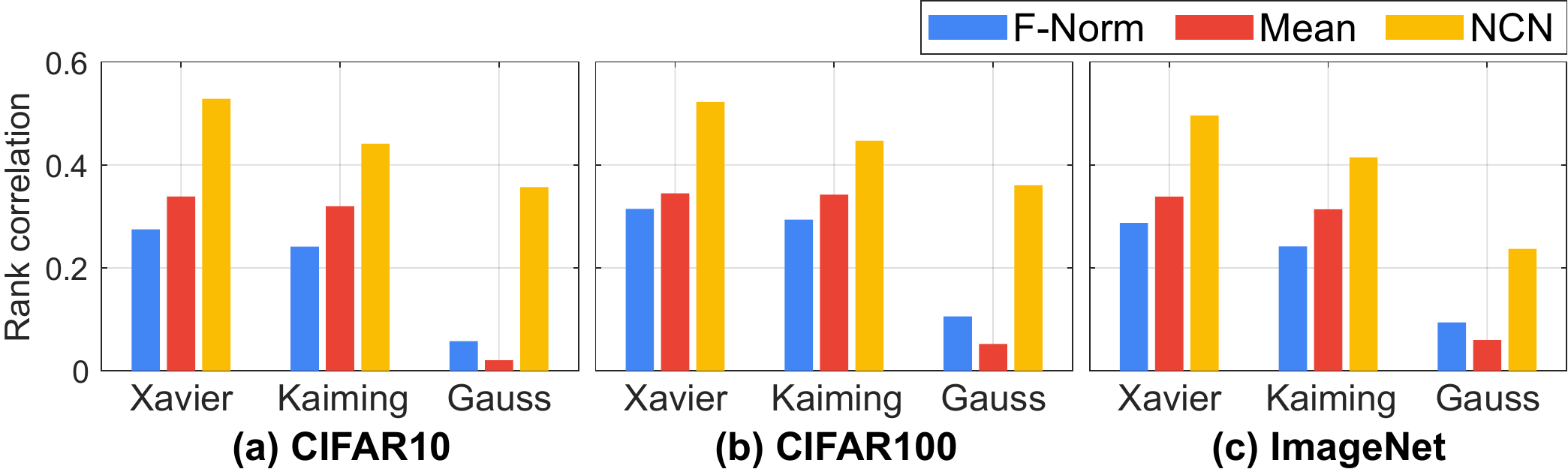}
\end{center}
\vspace{-10pt}
   \caption{ Rank correlation evaluation results on NAS-Bench-201 obtained from different initialization schemes. NCN appears relatively robust to change in initialization schemes, but F-Norm and Mean are destroyed with the Gaussian initialization.}
\label{fig:initialization}
\vspace{-15pt}
\end{figure}

Considering that previous NTK-based metrics have always been computed at initialization, the choice of weight initialization can be expected to have a non-negligible influence over the NTK computation result.
We test how the NTK-based metrics are affected by Xavier~\cite{glorot2010understanding}, Kaiming~\cite{he2015delving} and Gaussian initializations.
The experiments in this section are conducted only on NAS-Bench-201 as well.
~\figurename~\ref{fig:initialization} shows the change in rank correlation according to different initialization schemes.
All three metrics show some degree of fluctuation when using Xavier and Kaiming initializations, but when the Gaussian initialization is used, to our surprise, the rank correlation for F-Norm and Mean plummet close to zero.
This is an unexpected result because the NTK framework assumes that the parameters in a DNN are initialized as iid Gaussians, and thus their function realizations asymptotically converge to a Gaussian distribution in the infinite width limit~\cite{jacot2018neural}.

\section{Methodology}
We conjecture that the unreliability of the NTK-based metrics obtained at initialization occurs because the underlying theoretical assumptions in the NTK framework are violated in modern DNNs.
As a result, the NTK derived from a modern DNN is likely to evolve in a non-linear manner as training progresses, diverging away from the NTK at initialization~\cite{goldblum2019truth, fort2020deep, ortiz2021can}.
In Section~\ref{time_evo_ntk}, we witness that the architectures considered in NAS indeed exhibit highly non-linear characteristics.
Then, in Section~\ref{lga}, we present Label-Gradient Alignment, a novel NTK-based metric that has yet to be studied in NAS, and show how it can capture the evolution of the NTK with respect to target labels.
Afterwards, in Section~\ref{time_evo_metrics}, we corroborate the theoretical motivation behind LGA by demonstrating that after little amount of training, LGA shows a meaningful level of rank correlation with the test accuracy.
Please refer to the Appendix for detailed experimental settings used in this section.

\subsection{Time Evolution of the NTK}\label{time_evo_ntk}
\begin{figure}[t]
\begin{center}
   \includegraphics[width=\linewidth]{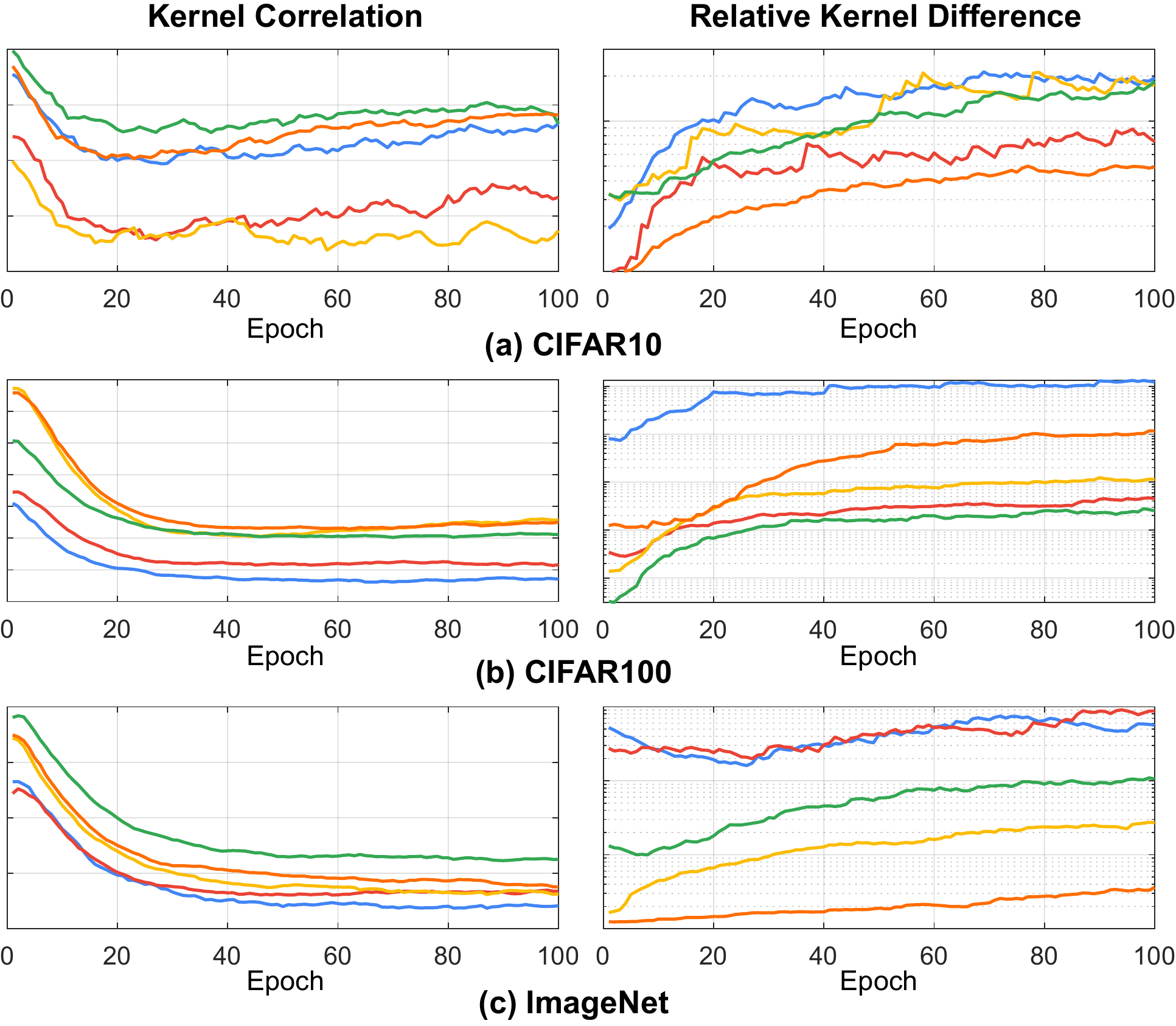}
\end{center}
\vspace{-10pt}
   \caption{ Analyzing the time evolution of the NTK as training progresses. Five unique architectures, represented by lines of different colors, are randomly sampled. For all three datasets, the kernel correlation decreases, and the kernel difference increases.}
\label{fig:time_evo_analysis}
\vspace{-10pt}
\end{figure}

\textbf{Kernel Correlation} measures the Pearson's correlation coefficient between $\Theta_{\theta_0}$ and $\Theta_{\theta_t}$: 
 $\mathrm{Cov} \left(\Theta_{\theta_0}, \Theta_{\theta_t} \right) \slash (\sigma \left(\Theta_{\theta_0} \right) \sigma \left(\Theta_{\theta_t} \right))$.
The correlation measurement results are presented on the left panel of~\figurename~\ref{fig:time_evo_analysis}.
For all three datasets, the correlation between $\Theta_{\theta_0}$ and $\Theta_{\theta_t}$ decreases rapidly in the initial epochs and start to stabilize after some amount of training, and such a trend becomes more conspicuous with the growth in data complexity. 

\textbf{Relative Kernel Difference} measures the relative change in the NTK from $\Theta_{\theta_0}$ to $\Theta_{\theta_t}$: $| \Theta_{\theta_t} - \Theta_{\theta_0} | \slash |\Theta_{\theta_0} |$.
The kernel difference measurement results are visualized on the right panel of~\figurename~\ref{fig:time_evo_analysis}.
We once again observe that the NTK deviates noticeably from $\Theta_{\theta_0}$ in the initial epochs, but the relative difference starts to saturate mid-training.

Based on both the correlation and the distance measurement results, a singular conclusion can be drawn: modern neural architectures primarily studied in NAS exhibit a highly non-linear behavior during training, and thus, the NTK in such architectures experiences a large amount of kernel twisting.
Consequently, the NTK framework, whose core theoretical results are built on the assumption that the NTK remains constant throughout training, loses its credibility, and the characteristics of the NTK at initialization become incapable of accurately representing the final test accuracy of a neural architecture.
This finding can be interpreted as being consistent with recent discovery that the non-linear advantage in DNNs is what allows them to outperform their linear kernel counterparts~\cite{fort2020deep, ortiz2021can}.
Therefore, we introduce Label-Gradient Alignment, a novel NTK-based metric that can capture the non-linear characteristics of neural architectures with only few epochs of training.

\subsection{Label-Gradient Alignment}\label{lga}
While a neural architecture's generalization performance must be one of the most prioritized factors in NAS, obtaining a closed-form characterization of generalization error on test data that cannot be accessed is impossible.
However, provided that the approximation in Eq.~(\ref{linear_approx}) is exact, it is possible to formulate generalization guarantees for DNNs by transferring the generalization bounds computed from their linear kernel equivalents.
In Bartlett~\textit{et al.}~\cite{bartlett2002rademacher}, it is shown that with high probability, the following relationship holds:
\begin{equation}\label{gen_bound}
    \mathcal{R}(f^*) \leq \hat{\mathcal{R}}(f^*) + \mathcal{O}\left(\sqrt{\frac{||f||^2_{\Theta_{\theta_0}}\mathbb{E}_x[\Theta_{\theta_0}(x, x')]}{m}}\right),
\end{equation}
where $\mathcal{R}$ and $\hat{\mathcal{R}}$ denote the expected and the empirical risks, respectively, of $f^* = \mathrm{argmin}_{h \in \mathcal{H}} \hat{\mathcal{R}}(h) + 
r||h||^2_{\Theta_{\theta_0}}$, with $r > 0$ as a regularization constant.
$f: \mathbb{R}^d \rightarrow \pm 1$ corresponds to the target function that the DNN is trying to learn, $||f||_{\Theta_{\theta_0}}$ is the RKHS norm of this target function.
One can assume that a DNN generalizes well whenever it achieves a low expected risk, and Eq.~(\ref{gen_bound}) indicates that the difference between the expected and the empirical risks decreases as the term $||f||^2_{\Theta_{\theta_0}}$ becomes smaller.
Through eigendecomposition, $||f||^2_{\Theta_{\theta_0}}$ can be re-written as:
\begin{equation}\label{decomp}
    ||f||^2_{\Theta_{\theta_0}} = \sum_k \frac{1}{\lambda_k} (\mathbb{E}_{x\sim \mathcal{D}}[v_k(x)f(x)])^2,
\end{equation}
where $\{\lambda_k, v_k\}$ denotes $k$-th eigenvalue-eigenvector pair of $\Theta_{\theta_0}$.
We can now see that better generalization performance may be expected when targets (or labels) align well with top eigenvectors of the NTK matrix, \ie the first principal components of the variability of the per-sample gradients.

Instead of directly computing Eq.~(\ref{decomp}), Ortiz~\textit{et al.}~\cite{ortiz2021can} offer a more tractable bound on $||f||^2_{\Theta_{\theta_0}}$:
\begin{equation}\label{ortiz_bound}
\begin{aligned}
    & ||f||^4_2~\slash~||f||^2_{\Theta_{\theta_0}} \leq \alpha(f), \\
    & \mathrm{where}~\alpha(f) = \mathbb{E}_{x, x' \sim \mathcal{D}} [f(x) \Theta_{\theta_0}(x, x') f(x')],
\end{aligned}
\end{equation}
where $||\cdot||_{\Theta_{\theta_0}}$ and $||\cdot||_2$ denote the RKHS norm and the $l_2$ norm, respectively.
In a fully supervised setting, where the target function is defined in the form of class labels, target function $f$ in $\alpha(f)$ can be replaced with target labels $\mathcal{Y}$, thereby yielding:
\begin{equation}
    \alpha(\mathcal{Y}) = \mathcal{Y}^\intercal \Theta_{\theta_0} \mathcal{Y}.
\end{equation}
From here on, we refer to $\mathcal{Y}^\intercal \Theta_{\theta_0} \mathcal{Y}$ as \textit{LGA}, a shorthand for label-gradient alignment.
Replacing $\alpha(f)$ in Eq.~(\ref{ortiz_bound}) with $\alpha(\mathcal{Y})$, we see that smaller $||f||^2_{\Theta_{\theta_0}}$ will increase $\alpha(\mathcal{Y})$.
From Eq.~(\ref{gen_bound}), it is evident that a small value of $||f||^2_{\Theta_{\theta_0}}$ is preferred to minimize the gap between the expected and the empirical risks.
Therefore, LGA must be \textit{positively correlated} with the generalization performance of an architecture and thus with its final test accuracy.
The noteworthy difference between LGA and previously-proposed metrics is that LGA takes both the NTK and the target labels into consideration.
Such a mathematical formulation of LGA allows it to accurately follow the orientation of the NTK with respect to the target functions that the neural architecture is trying to estimate.

A similar measure was used in Deshpande~\textit{et al.}~\cite{deshpande2021linearized} in the context of model selection for finetuning.
Inspired by Deshpande~\textit{et al.}, we introduce additional procedures to effectively utilize LGA for NAS.
To extend the binary classification setting of the NTK framework to multi-class classification in NAS, an $N \times N$ label matrix $L_\mathcal{Y}$, is introduced, in which $L_\mathcal{Y}[i, j] = 1$ if $x_i$ and $x_j$ belong in the same class, and $L_\mathcal{Y}[i, j] = -1$ otherwise. 
To induce the invariability in scale, LGA is normalized as follows:
\begin{equation}
    LGA = \frac{(\Theta_{\theta_0} - \mu(\Theta_{\theta_0})) \cdot (L_\mathcal{Y} - \mu(L_\mathcal{Y}))}{||\Theta_{\theta_0} - \mu(\Theta_{\theta_0})||_2 ||L_\mathcal{Y} - \mu(L_\mathcal{Y})||_2}, 
\end{equation}
where $\mu(L_\mathcal{Y})$ is the average of elements in $L_\mathcal{Y}$.

\subsection{NTK-based Metrics after Training}\label{time_evo_metrics}
\begin{figure}
\begin{center}
   \includegraphics[width=\linewidth]{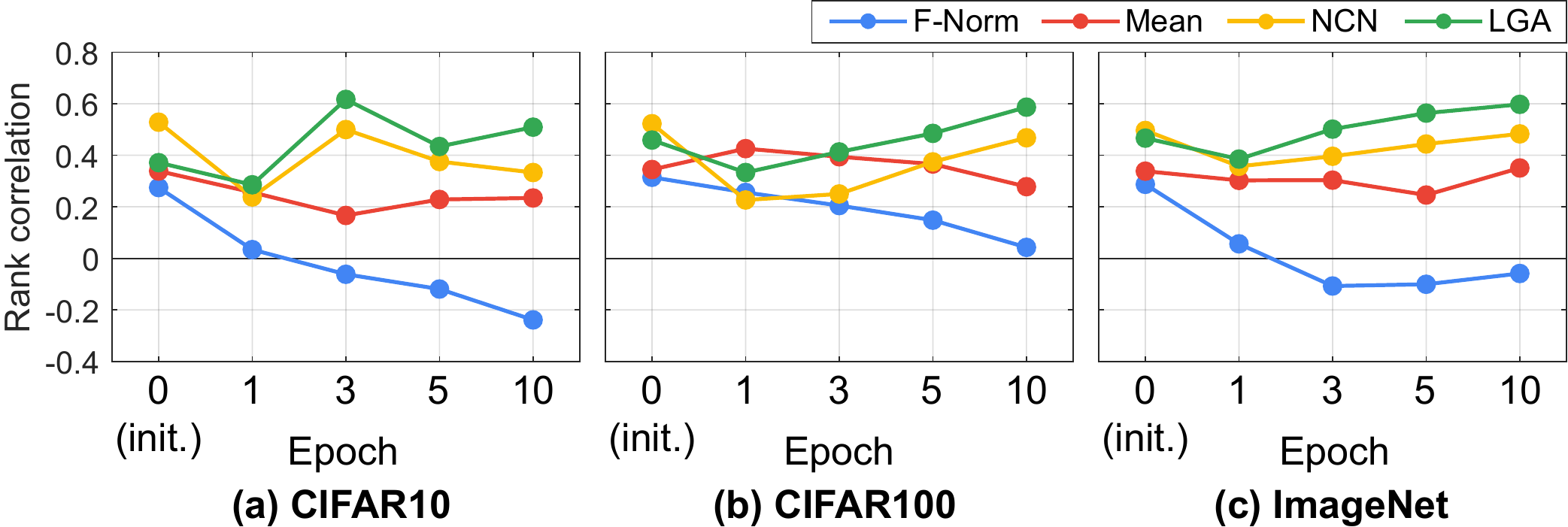}
\end{center}
\vspace{-10pt}
   \caption{Post-training rank correlation evaluation results on NAS-Bench-201. Regardless of the dataset, the predictive performance of LGA steadily improves from initialization. }
\label{fig:with_train}
\vspace{-10pt}
\end{figure}

We now repeat the rank correlation evaluation experiment conducted in Section~\ref{sec:3} after training architectures for $t \in \{1, 3, 5, 10\}$ epochs.
The post-training rank correlation evaluation results on NAS-Bench-201 are visualized in~\figurename~\ref{fig:with_train}.
In terms of rank correlation, LGA is the only metric that exhibits a steady improvement as training progresses across all three datasets.
In~\tablename~\ref{table:post_train_other_benchmarks}, we compare LGA after a single training epoch (LGA$_1$) with NTK-based metrics obtained at initialization on other NAS benchmarks.
Independent of the choice of a benchmark, the rank correlation of LGA$_1$ exceeds that of previous NTK-based metrics.

To better understand this characteristic behavior of LGA, we analyze how the NTK-based metrics of high-, mid-, and low-accuracy architectures change during the training process; please refer back to~\figurename~\ref{fig:metric_bench_201_all} for the results of this analysis.
For high-accuracy architectures, we observe a surge in LGA, which indicates that the concentration of labels on to the principal components of the NTK matrix has significantly increased; LGA in mid-accuracy architectures behaves similarly in that it shows some amount of increase, albeit small. 
On the contrary, for low-accuracy architectures, LGA remains stationary.
While the other metrics also change during training, they do so in a rather meaningless way that cannot distinguish different amounts of non-linear advantage in high-, mid- and low-accuracy architectures.
This experimental analysis justifies the need for target labels in LGA to understand how the NTK rotates and evolves with respect to the target functions.
As stated in Section~\ref{lga}, with target labels as a type of an anchor point, LGA can distinguish between highly trainable architectures, in which target labels gradually become more aligned with the principal components of the NTK in the initial epochs, and less trainable architectures, which do not benefit much from kernel twisting.
In the absence of target labels, the other metrics cannot determine the direction in which the NTK evolves, and hence, they do not gain any meaningful information from the training process.

\begin{table}[t]
\centering
\caption{Comparison of LGA$_1$ with previous NTK-based metrics on various NAS benchmarks. In terms of rank correlation, LGA$_1$ outperforms other metrics across all benchmarks.}
\vspace{-5pt}
\setlength{\tabcolsep}{6pt}
\renewcommand{\arraystretch}{0.9}
\begin{tabular}{l|rrr|r}
\toprule
Benchmark & F-Norm & Mean & NCN & LGA$_1$ \\
\bottomrule\toprule
NAS-Bench-101 & -0.022 & -0.023 & 0.094 & {\bf 0.308} \\
NDS-DARTS & 0.234 & 0.062 & 0.145 &  {\bf0.408} \\
NDS-ENAS & 0.065 & 0.037 & 0.162 & {\bf0.416} \\ 
NDS-NASNet & -0.073 & -0.075 & 0.146 & {\bf0.357} \\
NDS-Amoeba & 0.157 & 0.158 & 0.013 & {\bf0.396} \\
\bottomrule
\end{tabular}
\vspace{-10pt}
\label{table:post_train_other_benchmarks}
\end{table}

\input{tex/nb201_search}

\section{Searching with LGA}\label{lga_search}
To demonstrate that LGA can be utilized to improve the computational efficiency of NAS, we integrate LGA with random search and evolutionary search algorithms.
Based on the evaluation results in Section~\ref{time_evo_ntk}, LGAs after 3 (LGA$_3$) and 5 (LGA$_5$) epochs of training are used for searching.
In random search (RS), 100 architectures are sampled from the search space for evaluation, and the architecture with the maximum LGA is selected.
For evolutionary search (REA), we adopt the regularized evolutionary search algorithm of Real~\textit{et al.}~\cite{real2019regularized}.
The regularized approach of Real~\textit{et al.} differs from a na\"ive evolutionary algorithm in that it prefers newer candidate architectures.
We use a fixed search cost budget for evolutionary search on all three datasets. 
Please refer to the Appendix for the experimental settings used in this section and comprehensive pseudo codes of both search processes.

In \tablename~\ref{table:search_result}, the search performances of RS and REA with LGA$_3$ and LGA$_5$ are compared against those of other state-of-the-art NAS algorithms.
We would like to emphasize that for both RS and REA, no additional information besides LGA is used to evaluate architectures.
RS with either LGA$_3$ or LGA$_5$ outperforms search algorithms based on other NTK-based metrics obtained at initialization; TE-NAS~\cite{chen2020neural} and KNAS~\cite{xu2021knas} utilize CN and Mean, respectively. 
This result is particularly impressive considering that TE-NAS and KNAS rely on some external signal during search; TE-NAS utilizes a more complicated search algorithm and another at-initialization metric, and in the final architecture derivation step of KNAS, the actual test accuracy is used.
REA with LGA$_3$ or LGA$_5$ also achieves competitive test accuracy, and more importantly, it does so with far less search cost than other search algorithms on CIFAR-100 and ImageNet-16-120.
Overall, despite introducing some amount of training, LGA$_3$ and LGA$_5$ appear to be highly competent and computationally efficient metrics.
Lastly, we show that LGA can be applied more broadly to a vareity of search spaces by conducting random search on various bechmarks other than NAS-Bench-201.
The results and experimental details are provided in the Appendix.

\section{Concluding Remarks}
The technical and experimental contributions of this paper are largely three-fold.
First, through a more extensive and fine-grained evaluation of the NTK-based metrics, we revealed that the current form of the NTK framework might not be as reliable of a theoretical framework for NAS as previously believed to be.
Second, through the empirical analysis of the time evolution of the NTK, we demonstrated that the aforementioned limitation occurs because in modern neural architectures, the NTK evolves in a highly non-linear manner during the training process, diverging significantly away from the NTK at initialization.
Third, when complemented with little amount of training, LGA, first introduced in this work, rose as a strong predictor of test accuracy because its innate theoretical motivation could embody the non-linear characteristics of modern neural architectures, which the other NTK-based metrics were blind to. 
Integrating LGA into existing search algorithms provided a further empirical support for its effectiveness as a computationally effective predictor of test accuracy.
We discuss the limitations and societal impact of our work in the Appendix.
\section*{Acknowledgements}
This work was supported by Institute of Information \& communications Technology Planning \& Evaluation (IITP) grant funded by the Korea government(MSIT) [NO.2021-0-01343, Artificial Intelligence Graduate School Program (Seoul National University)], the BK21 FOUR program of the Education and Research Program for Future ICT Pioneers, Seoul National University in 2022, AIRS Company in Hyundai Motor Company \& Kia Corporation through HMC/KIA-SNU AI Consortium Fund, and SNU-Naver Hyperscale AI Center.



\clearpage



\section*{Appendices}
\setcounter{section}{0}
\renewcommand\thesection{A\arabic{section}}
\setcounter{table}{0}
\renewcommand{\thetable}{A\arabic{table}}
\setcounter{figure}{0}
\renewcommand{\thefigure}{A\arabic{figure}}
\setcounter{equation}{0}
\renewcommand{\theequation}{A\arabic{equation}}

\section{Related Works}
\subsection{Neural Architecture Search}
In early NAS, evolutionary algorithm (EA)~\cite{real2017large, real2019regularized} or reinforcement learning (RL)~\cite{le2017naswithRL, zoph2018nasnet, raskar2016metaqnn} were commonly-used search algorithms.
Although these early works were successful in proving the potential of NAS, their immediate application was challenging due to the enormous search cost, easily mounting up to tens of thousands of GPU hours to search for a single architecture.
Most of the computational overhead in early NAS algorithms occurred as the result of having to train each candidate architecture to convergence and evaluate it.

Through weight sharing~\cite{dean2018enas}, recent NAS works were able to achieve a noticeable acceleration in the architecture search process.
Weight sharing utilizes a super-network, whose sub-networks correspond to candidate architectures belonging in the pre-defined search space.
The sub-networks are evaluated with the weights inherited from the super-network, and thus, the sub-networks end up sharing a common set of weights.
Modern NAS algorithms that exploit such performance approximation techniques can be categorized into differentiable NAS~\cite{yang2019darts, chen2020stabilizing, xu2019pc, zela2019understanding, chu2019fairdarts, chen2019progressive, li2020sgas, zhou2020NAS, chu2020darts, chen2020drnas}, and one-shot NAS~\cite{le2018oneshot, dong2019searching, brock2018smash, peng2020cream, zhang2020overcoming}; while the search and evaluation processes are entangled in the former, the latter disentangles them into separate processes.

Performance predictors~\cite{deng2017peephole}, which take an encoded neural architecture as an input and output the accuracy of the corresponding architecture, are another promising direction for reducing the architecture evaluation cost.
The main challenge in performance prediction is minimizing the number of architecture-accuracy pairs required to obtain a performance predictor that generalizes well to the rest of the search space.
Because the problem of architecture performance prediction is by definition a regression task,~\cite{deng2017peephole, tang2020semi, li2020gp} aimed to predict the exact value of accuracy by minimizing the mean squared error loss between the predicted and true accuracy. 
Recently, the concept of architecture comparators~\cite{xu2021renas, chen2021contrastive}, is rising in popularity.
Instead of estimating the exact accuracy, comparators take two architectures as an input and use a ranking loss or a contrastive learning framework to predict which is more likely to rank higher in terms of accuracy.

Apart from weight sharing and performance predictors, there also exist works that aim to search for more general proxy settings for architecture evaluation.
EcoNAS~\cite{zhou2020econas} explores four common reduction factors - the number of channels, the resolution of input images, the number of training epochs, and the sample ratio of the full training set - and determine which one of these proxies can be used to reliably estimate the final test accuracy.
Na~\textit{et al.}~\cite{na2021accelerating} show that it is possible to only use a subset of the target dataset for execute NAS and propose a novel proxy dataset selection algorithm.

\subsection{NAS at initialization}
Evaluating neural architectures without any amount of training is surely an interesting and attractive research direction that has potential to NAS.
Mellor~\textit{et al}.~\cite{mellor2021neural} use the feature separability in the linear regions of a neural architecture as a metric to score architectures.
Abdelfattah~\textit{et al.}~\cite{abdelfattah2020zero} attempt to identify which one of the pruning-at-initialization techniques is most useful for NAS.
Zen-NAS~\cite{lin2021zen} analyzes the activation patterns in a neural architecture to quantify its expressivity.
KNAS~\cite{xu2021knas} and TE-NAS~\cite{chen2020neural} have previously proposed to use the NTK framework to score neural architectures and thus are most closely-related to this work.
As mentioned in the main paper, KNAS and TE-NAS use the MEAN and the CN metrics, respectively.
In addition to CN, TE-NAS utilizes another at-initialization score, derived from the number of linear regions in a neural architecture. 

\subsection{Neural Tangent Kernel}
The NTK framework is based on the observation that for certain initialization schemes, the infinite width limit of many neural architectures can be exactly characterized using kernel tools~\cite{jacot2018neural}. 
Provided that this assumption holds, many of the questions in deep learning theory can be addressed through the study of linear methods and convex analyses~\cite{scholkopf2002learning}.
The intuitiveness of the NTK framework led to important results regarding the generalization and optimization of deep neural networks~\cite{lee2019wide, du2019gradient, arora2019exact, bietti2019inductive, zou2019improved, liu2020linearity}.
Such advances in the NTK framework subsequently led researchers to study how the NTK can be leveraged in various applications: prediction of the generalization performance and the training speed, explanation of inductive biases in deep neural networks, design of new classifiers, 
Despite these limitations, the intuitiveness of the NTK, which allows to use a powerful set of theoretical tools to exploit it, has led to a rapid increase in the amount of research that successfully leverages the NTK in applications, such as predicting generalization~\cite{deshpande2021linearized} and training speed~\cite{zancato2020predicting}, explaining certain inductive biases~\cite{mobahi2020self, tancik2020fourier} or designing new classifiers~\cite{arora2019harnessing, maddox2021fast}.
Despite the proliferation of the NTK framework in the deep learning theory, there still exist doubts on whether the assumptions in the NTK framework truly holds for deep neural networks that are used in real life~\cite{goldblum2019truth, fort2020deep}.

\begin{table}[t]
\centering
\caption{Overview of the NTK-based metrics studied in the main paper. ``Rank" refers to whether the metric should be positively ($+$) or negatively ($-$) correlated with the final test accuracy.}
\setlength{\tabcolsep}{5pt}
\renewcommand{\arraystretch}{1.5}
\begin{tabular}{l|c|c}
\toprule
Metric & Equation & Rank \\
\bottomrule\toprule
F-Norm &  $||\Theta_{\theta_t}||_F$ & $+$ \\
Mean & $\mu({\Theta}_{\theta_0})$ & $+$ \\
NCN & $\frac{-(\lambda_\mathrm{max}({\Theta}_{\theta_0}))}{(\lambda_\mathrm{min}({\Theta}_{\theta_0}))}$ & $+$ \\
LGA & $\frac{(\Theta_{\theta_0} - \mu(\Theta_{\theta_0})) \cdot (L_\mathcal{Y} - \mu(L_\mathcal{Y}))}{||\Theta_{\theta_0} - \mu(\Theta_{\theta_0})||_2 ||L_\mathcal{Y} - \mu(L_\mathcal{Y})||_2}$ & $+$\\
\bottomrule
\end{tabular}
\label{table:metric_summary}
\end{table}

\begin{table}[t]
\centering
\caption{Summary of differences among different NDS search spaces. ``\# Ops." and ``\# nodes" correspond to the number of opearations and the number of nodes in each cell. ``Output" refers to which node(s) are concatenated for the output ($=$ A if all nodes are are concatenate, $=$ L if there are nodes that are not used as input to other nodes). ``\# cells" refers to the number possible cells, without considering the redundancy, that exist in each search space.}
\setlength{\tabcolsep}{5pt}
\renewcommand{\arraystretch}{0.9}
\begin{tabular}{l|ccc|r}
\toprule
Benchmarks & \# Ops. & \# Nodes & Output & \# Cells (B) \\
\bottomrule\toprule
NASNet & 13 & 5 & L & 71,465,842 \\
Amoeba & 8 & 5 & L & 556,628\\
PNAS & 8 & 5 & A & 556,628 \\
ENAS & 5 & 5 & L & 5,063 \\
DARTS & 8 & 4 & A & 242 \\
\bottomrule
\end{tabular}
\label{table:nds_summary}
\end{table}

\section{Metrics Summary} In~\tablename~\ref{table:metric_summary}, we provide an overview of the NTK-based metrics studied in the main paper, along with the direction of their rank correlation with the final test accuracy of a neural architecture. 

\section{NAS Benchmarks \& Image Datasets}
\noindent\textbf{NAS-Bench-101} (cont'd from the main paper) Each convolution operator in NAS-Bench-101 follows the Conv-BN-ReLU pattern, and na\"ive convolutions are used instead of separable convolutions, such that the resulting architectures closely match the designs of ResNet and Inception. Each cell is stacked 3 times, followed by a max-pooling layer, which halves the image height and width and doubles the number of channels. In the resulting DNN, the above pattern is repeated 3 times, and lastly, a glabal average pooling and a final classification layer with the Softmax function are inserted. \\
\noindent\textbf{NAS-Bench-201} (cont'd from the main paper) The convolution operator in NAS-Bench-201 follows the operation sequence of ReLU-Conv-BN. The macro-architecture of NAS-Bench-201 starts with one 3-by-3 convolution with 16 output channels and a batch normalization layer~\cite{ioffe2015batch}. Each cell is stacked 5 times, followed by a residual block. In the final DNN, this pattern is repeated 3 times. The number of output channels in the first, second, and third stages is set to be 16, 32 and 64, respectively. The residual block serves to downsample the spatial size and double the channels of an input feature map. The shortcut path in this residual block consists of a 2-by-2 average pooling layer with stride of 2 and a 1-by-1 convolution. Lastly, a global average pooling and a final classification layer with the Softmax function are inserted for classification.
\\
\noindent\textbf{NDS Benchmark}~\cite{radosavovic2019network} Please refer to~\tablename~\ref{table:nds_summary} for the summary of differences among the search spaces in the NDS benchmark~\cite{radosavovic2019network} \\
\noindent\textbf{Image Datasets} Please refer to~\tablename~\ref{table:dataset_config} for the summary of image datasets utilized in NAS benchmarks. 

\section{Experimental Details}
\noindent\textbf{Section 3.2} The NTK computation involves per-sample gradients, which are computationally intractable to obtain from high-dimensional datasets that contain tens of thousands images. Therefore, in this paper, we instead use a single minibatch, randomly sampled from the train set, to compute the NTK. For CIFAR-10 and CIFAR-100 datasets, a minibatch of size 256 is used, while for ImageNet16-120, a minibatch of size 512 is used. For a fair comparison, we use the same set of image samples to construct the minibatch used for evaluation across all benchmarks. A single NVIDIA V100 GPU is used for the experiments in this section. \\
\noindent\textbf{Sections 3.3. \& 3.4} For rank correlation evaluation on CIFAR-10 and CIFAR-100, we use a minibatch of size 256, while for that on ImageNet16-120, we use a minibatch of size 512. For NCN, we use the Eval mode BN on PyTorch, and for F-Norm and Mean, we use the Train mode BN. The choice of BN usage is set based on the results from Section 3.2; for all metrics, we choose the BN setting that yields the highest rank correlation for each metric. A single NVIDIA V100 GPU is used for the experiments in these sections.  \\
\noindent\textbf{Section 4.1} Measurements are done on NAS-Bench-201. We use a single image sampled from the validation set and Eval mode BN for NTK computation in this section. A single NVIDIA V100 GPU is used for the experiments. \\
\noindent\textbf{Section 4.3} To train candidate architectures, we use the momentum SGD optimizer with a learning rate of 0.025, momentum of 0.9, a weight decay factor of 3e-4. These are standard settings used to train architectures in NAS benchmarks. The train set of each dataset is used for training, and a single minibatch from the validation set is used to derive NTK-based metrics. A batch size of 1024 is used to train architectures. For NTK computation, a batch size of 256 is used for CIFAR-10 and CIFAR-100, and a batch size of 512 is used for ImageNet16-120. A single NVIDIA V100 GPU is used for the experiments in this section. For F-Norm and Mean, we use the Train mode BN, and for NCN and LGA, we use the Eval mode BN.\\
\noindent\textbf{Section 5} We use the same experimental settings as those used in~\textbf{Section 4.3} to train architectures and derive LGA$_3$ and LGA$_5$. A single NVIDIA A40 GPU is used to execute both random and evolutionary search algorithms.\\

\section{Full Benchmark Evaluation Results}
\begin{figure*}[t]
\begin{center}
  \includegraphics[width=\linewidth]{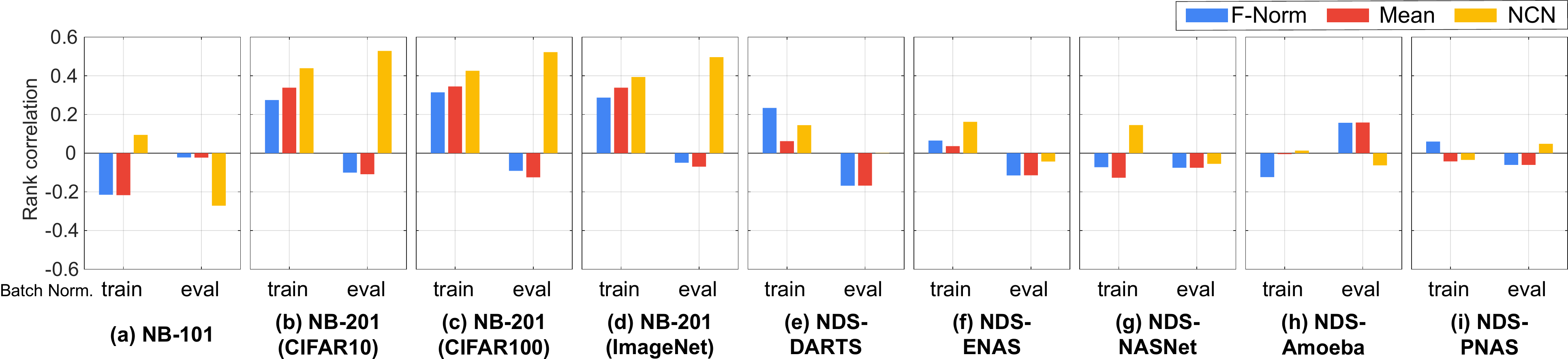}
\end{center}
\vspace{-10pt}
  \caption{Rank correlation evaluation results on various NAS benchmarks. For F-Norm and Mean, the evaluation results based on the Train mode BN are reported, whereas for NCN and LGA, those based on the Eval mode BN are reported. The scale and the range of y-axes are set to be the same across all search spaces. NB is an abbreviation for NAS-Bench.}
\label{fig:nasbench}
\vspace{-5pt}
\end{figure*}

In~\figurename~\ref{fig:nasbench}, the rank correlation evaluation results of existing at-initialization NTK-based metrics on all benchmarks are visualized.

\section{Fine-grained Rank Correlation Evaluation}
\begin{figure*}[t]
\begin{center}
  \includegraphics[width=0.8\linewidth]{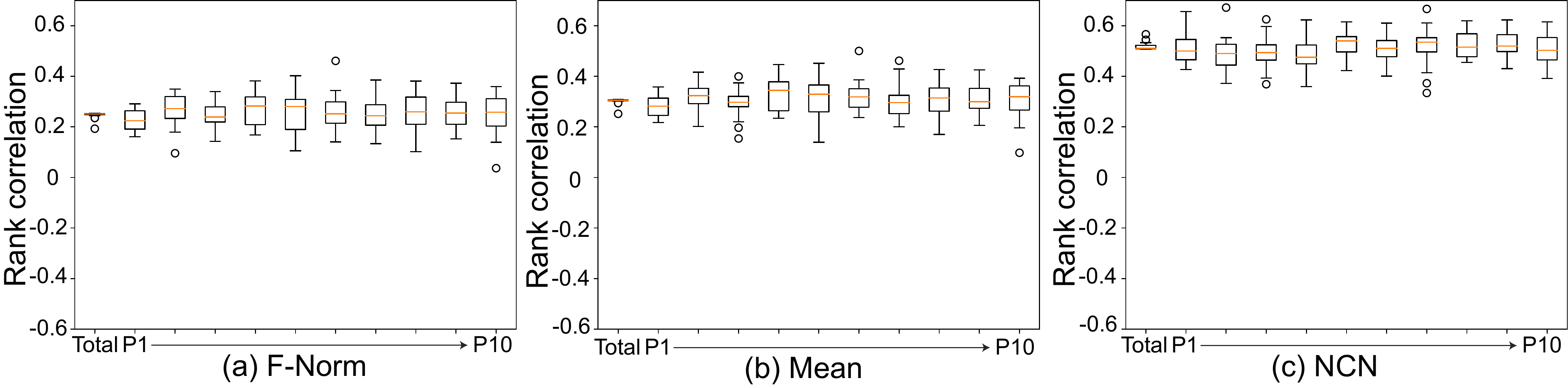}
\end{center}
\vspace{-10pt}
  \caption{Rank correlation evaluation results on NAS-Bench-201 CIFAR-10 for every accuracy decile.}
\label{fig:c10_percentile}
\end{figure*}

\begin{figure*}[t]
\begin{center}
  \includegraphics[width=0.8\linewidth]{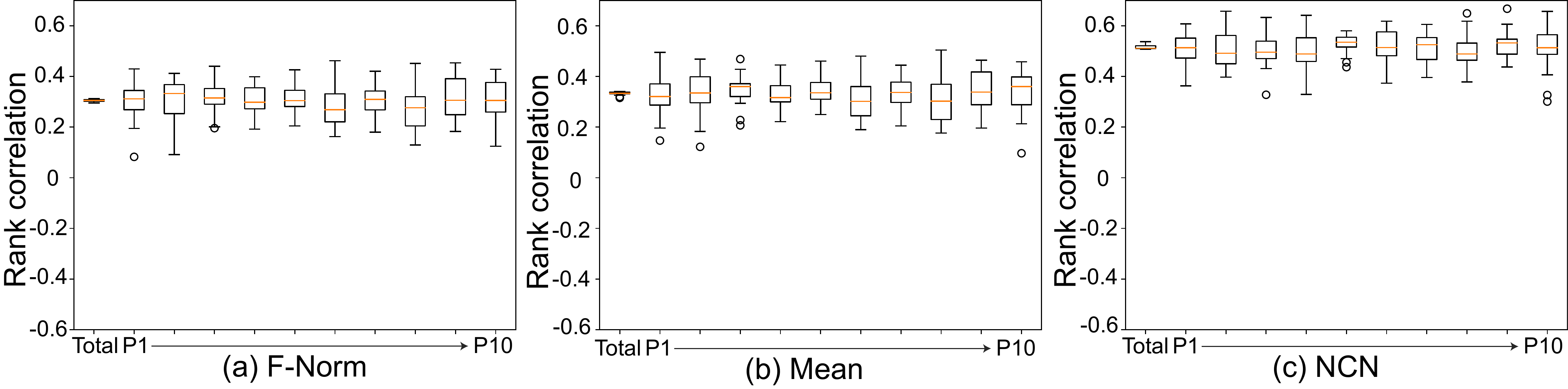}
\end{center}
\vspace{-10pt}
\caption{Rank correlation evaluation results on NAS-Bench-201 CIFAR-100 for every accuracy decile.}
\label{fig:c100_percentile}
\end{figure*}

\begin{figure*}[t]
\begin{center}
  \includegraphics[width=0.8\linewidth]{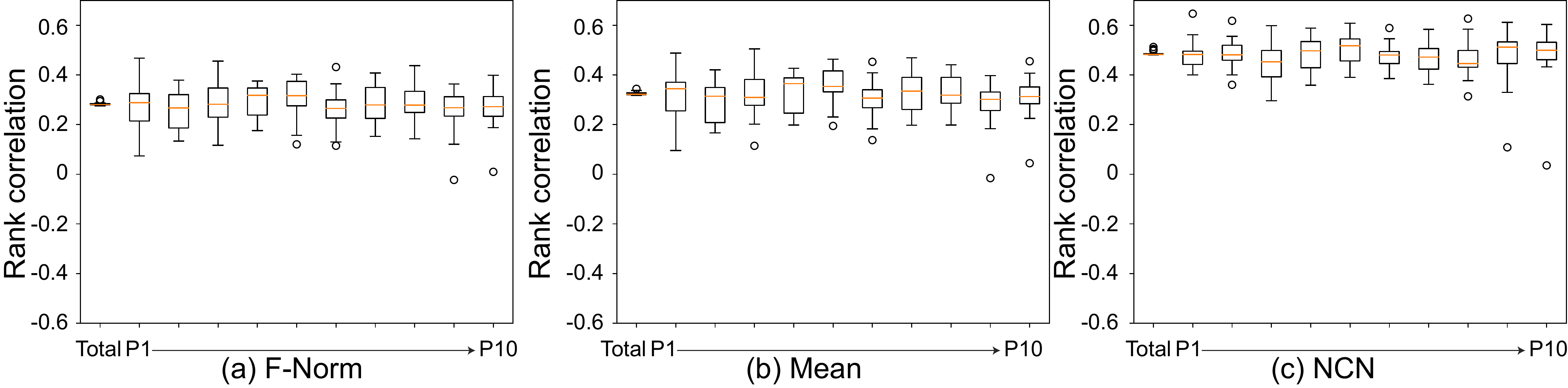}
\end{center}
\vspace{-10pt}
\caption{Rank correlation evaluation results on NAS-Bench-201 ImageNet-16-120 for every accuracy decile.}
\label{fig:imgnet_percentile}
\end{figure*}

This experiment is repeated for 20 different runs, and the evaluation results in box-and-whisker plots are presented in~\figurename~\ref{fig:c10_percentile},~\ref{fig:c100_percentile}, and~\ref{fig:imgnet_percentile}.
P1 contains 100 architectures sampled from Top-$10\%$ of architectures, whereas P10 contains the same number of architectures sampled from Bottom-$10\%$ of architectures. Ticks on the x-axis correspond to accuracy deciles in descending order (P1 $\rightarrow$ P10). ''Total" refers to the rank correlation evaluation results over all 1,000 architectures.

\begin{table}[t]
\small
\centering
\caption{CIFAR-10 Search results on additional search spaces.}
\vspace{-1em}
\setlength{\tabcolsep}{3pt}
\renewcommand{\arraystretch}{0.9}
\begin{tabular}{lccc}
\toprule
Metric & NB101 & NDS-DARTS & NDS-ENAS \\
\bottomrule
\toprule
F-Norm & 89.17 & 91.83 & 85.19 \\
Mean & 88.05 & 88.40 & 91.14  \\
NCN & 90.81 & 91.99 & 92.52 \\
LGA & \textbf{92.57} & \textbf{93.61} & \textbf{93.26} \\
\toprule
Metric & NDS-Amoeba & NDS-NASNet & NAS-Macro \\ 
\bottomrule
\toprule
F-Norm & 91.35 & 84.71 & 87.12\\
Mean & 90.89 & 89.89 & 87.25 \\
NCN & 89.14 & 87.21 & 90.62 \\
LGA & \textbf{94.35} & \textbf{94.30} & \textbf{92.24} \\
\bottomrule
\end{tabular}
\label{table:extended_space}
\end{table}

\section{Search Results on Other Benchmarks}
We conducted random search with 500 sampled architectures with four compared metrics. Except for LGA, which is obtained after 3 epochs, others are measured at initialization. The search results are presented in~\tablename~\ref{table:extended_space}. LGA is the only metric that searches for a successful architecture across all search spaces. 

\section{Pseudocode for Search Algorithms}
The pseudocode for random search is presented in Algorithm 1. 
In random search, N corresponds to the total number of candidate architectures evaluated during search.
In this work, we set N in random search to be 100.
The pseudocode for evolutionary search is presented in Algorithm 2.
In evolutionary search, N corresponds to the number of architectures kept in the parent pool (or population).
In this work, we set N in evolutionary search to be 10.

\section{Limitations \& Societal Impact}
\textbf{Limitations:} Even though we demonstrate that LGA exhibits a high predictive performance with only few epochs of training, extensive theoretical advances are required to fundamentally address the unreliability of the NTK. To become a truly reliable theoretical framework for architecture selection, the NTK must be able to encompass diverse operations, normalization types, and weight initializations.\\
\textbf{Societal Impact:} With the help of at-initialization metrics, NAS was able to greatly reduce its environmental cost. Unfortunately, the need for training to stabilize the NTK will inevitably increase the environmental cost of NAS.

\begin{table*}[t]
\centering
\caption{Summary of image datasets used to construct NAS Benchmarks.}
\vspace{-5pt}
\setlength{\tabcolsep}{4pt}
\renewcommand{\arraystretch}{1.1}
\begin{tabular}{l|ccccc}
\toprule
Dataset & \# of Train Data & \# of Validation Data & \# of Test Data & \# of Classes & Image Size \\
\bottomrule\toprule
CIFAR-10 & 50,000 & - & 10,000 & 10 & $(32 \times 32)$ \\
CIFAR-100 & 50,000 & - & 10,000 & 100 & $(32 \times 32)$ \\
ImageNet-16-120 & 151,700 & 3,000 & 3,000 & 120 & $(16 \times 16)$ \\
\bottomrule
\end{tabular}
\label{table:dataset_config}
\end{table*}

\begin{algorithm}[h]
\SetAlCapFnt{\small}
\DontPrintSemicolon

sampler = RandomSampler() \;
best\_arch, best\_LGA = None, 0 \;
\textbf{for} $i = 1:N$ \textbf{do} \;
~~~cand\_arch = sampler() \;
~~~\textbf{for} $\mathrm{Epoch} = 1:t$ \textbf{do} \;
~~~~~~Train(cand\_arch) \;
~~~\textbf{end for} \;
~~~LGA$_t$ = cand\_arch.LGA() \;
~~~\textbf{if} LGA$_t$ $>$ best\_LGA \textbf{then} \;
~~~~~~best\_arch, best\_LGA = cand\_arch, LGA$_t$ \;
~~~\textbf{end if}

\textbf{end for}

chosen\_arch = best\_arch

\caption{Random Search}
\label{alg:random_search}
\end{algorithm}

\begin{algorithm}[h]
\caption{Evolutionary Search}
\label{alg:evo_search}
\SetAlCapFnt{\small}
\DontPrintSemicolon
parent\_pool = [] \;
lga\_hist = [] \;
sampler = RandomSampler() \;
parent\_arch, child\_arch = None, None\;
\textbf{for} $i = 1:N$ \textbf{do} \;
~~~cand\_arch = sampler() \;
~~~parent\_pool.append(cand\_arch) \; 
~~~\textbf{for} $\mathrm{Epoch} = 1:t$ \textbf{do} \;
~~~~~~Train(cand\_arch) \;
~~~\textbf{end for} \;
~~~lga\_hist.append(cand\_arch.LGA()) \;
\textbf{end for}

\BlankLine
\textbf{while} search budget not exceeded \textbf{do} \;
~~~Based on lga\_hist, choose the architecture with the highest LGA$_t$ from the parent pool as parent\_arch \;
~~~child\_arch = Mutate(parent\_arch) \;
~~~\textbf{for} $\mathrm{Epoch} = 1:t$ \textbf{do} \;
~~~~~~Train(child\_arch) \;
~~~\textbf{end for} \;
~~~parent\_pool.popleft() \;
~~~lga\_hist.popleft() \;
~~~parent\_pool.append(child\_arch) \;
~~~lga\_hist.append(child\_arch.LGA()) 

\end{algorithm}

\clearpage

{\small
\balance
\bibliographystyle{ieee_fullname}
\bibliography{main}
}

\end{document}

%% file: tex/nb201_search.tex
\begin{table*}[t]
\centering
\caption{ Comparison against state-of-the-art NAS algorithms on NAS-Bench-201. ``Optimal" refers to the best test accuracy achievable in the NAS-Bench-201 search space. The search process is executed separately for each image dataset. The search cost is reported in GPU seconds. All of our search experiments are conducted on a single NVIDIA Tesla A40 GPU. $^\dagger$: search based on NTK-based metrics.}
\setlength{\tabcolsep}{5pt}
\vspace{-3pt}
\renewcommand{\arraystretch}{0.9}
\begin{tabular}{l|ccc|ccc|ccc|c}
\toprule
& \multicolumn{3}{c|}{CIFAR-10} & \multicolumn{3}{c|}{CIFAR-100} & \multicolumn{3}{c|}{ImageNet-16-120} & Search \\
Model & Acc. & Cost & Speed-up & Acc. & Cost & Speed-up & Acc. & Cost & Speed-up & Method \\
\bottomrule\toprule
ResNet & 93.97 & N/A & N/A & 70.86 & N/A & N/A & 43.63 & N/A & N/A & Manual \\
\midrule
RS~\cite{bergstra2012random} & 93.63 & 216K & $1.0\times$ & 71.28 & 460K & $1.0\times$ & 44.88 & 1M & $1.0\times$ & Random \\
RL~\cite{le2017naswithRL} & 93.72 & 216K & $1.0\times$ & 70.71 & 460K & $1.0\times$ & 44.10 & 1M & $1.0\times$ & RL \\
REA~\cite{real2019regularized} & 93.72 & 216K & $1.0\times$ & 72.12 & 460K & $1.0\times$ & 45.01 & 1M & $1.0\times$ & EA \\
BOHB~\cite{falkner2018bohb} & 93.49 & 216K & $1.0\times$ & 70.84 & 460K & $1.0\times$ & 44.33 & 1M & $1.0\times$ & HPO \\
RSPS~\cite{li2020random} & 91.67 & 10K & $21.6\times$ & 57.99 & 46K & $21.6\times$ & 36.87 & 104K & $9.6\times$ & RS+WS \\
DARTS~\cite{yang2019darts} & 88.32 & 23K & $9.4\times$ & 67.34 & 80K & $5.8\times$ & 33.04 & 110K & $9.6\times$ & Gradient \\
GDAS~\cite{dong2019searching} & 93.36 & 22K & $12.0\times$ & 67.60 & 39K & $11.7\times$ & 37.97 & 130K & $7.7\times$ & Gradient \\
\midrule
NASWOT~\cite{mellor2021neural} & 92.96 & 2.2K & $100\times$ & 70.03 & 4.6K & $100\times$ & 44.43 & 10K & $100\times$ & Random \\
$^\dagger$TE-NAS & 93.90 & 2.2K & $100\times$ & 71.24 & 4.6K & $100\times$ & 42.38 & 10K & $100\times$ & Pruning-based \\
$^\dagger$KNAS ($k = 2$) & 93.05 & 4.2K & $50\times$ & 68.91 & 9.2K & $50\times$ & 34.11 & 20K & $50\times$ & Random \\
$^\dagger$KNAS ($k = 5$) & 93.42 & 10.8K & $20\times$ & 71.42 & 23K & $20\times$ & 45.35 & 50K & $20\times$ & Random \\
\midrule
$^\dagger$RS + LGA$_3$ & 93.64 & 3.6K & $60\times$ & 69.77 & 5K & $92\times$ & 45.03 & 10.1K & $99\times$ & Random \\
$^\dagger$RS + LGA$_5$ & 94.03 & 5.4K & $40\times$ & 71.56 & 7K & $66\times$ & \textbf{46.30} & 15K & $67\times$ & Random \\
$^\dagger$REA + LGA$_3$ & \textbf{94.30} & 3.6K & $60\times$ & 71.18 & 3.6K & $127\times$ & 45.30 & 3.6K & $277\times$ & EA \\
$^\dagger$REA + LGA$_5$ & 93.94 & 5.4K & $40\times$ & \textbf{72.42} & 5.4K & $85\times$& 45.17 & 5.4K & $185\times$ & EA \\
\midrule
Optimal & \multicolumn{3}{c|}{94.37} & \multicolumn{3}{c|}{73.51} & \multicolumn{3}{c|}{47.31} & N/A \\ 
\bottomrule
\end{tabular}
\vspace{-10pt}
\label{table:search_result}
\end{table*}